\documentclass[letterpaper]{article} 
\usepackage{aaai2026}  
\usepackage{times}  
\usepackage{helvet}  
\usepackage{courier}  
\usepackage[hyphens]{url}  
\usepackage{graphicx} 
\usepackage{natbib}  
\usepackage{caption} 
\usepackage{algorithm}
\usepackage{algorithmic}
\usepackage{footmisc}
\usepackage{marvosym}

\usepackage{newfloat}
\usepackage{listings}
\DeclareCaptionStyle{ruled}{labelfont=normalfont,labelsep=colon,strut=off} 
\floatstyle{ruled}
\newfloat{listing}{tb}{lst}{}
\floatname{listing}{Listing}
\usepackage{graphicx}
\usepackage{makecell}
\usepackage{color}
\usepackage{tabularray}
\usepackage{subcaption}
\usepackage{algorithm}
\usepackage{algorithmic}
\usepackage{tabularx}
\usepackage{tabularray}
\usepackage{amsmath}      
\usepackage{amssymb}      

\usepackage{xspace}
\usepackage{multirow}
\newcommand{\eat}[1]{}

\newcommand{\eg}{\emph{e.g.},\xspace}

\newcommand{\M}{SSPO\xspace}
\newcommand{\ML}{Self-traced Step-wise Preference Optimization\xspace}

\definecolor{Jade}{rgb}{0,0.69,0.313}
\definecolor{GuardsmanRed}{rgb}{0.752,0,0}

\nocopyright 

\title{SSPO: Self-traced Step-wise Preference Optimization for Process Supervision and Reasoning Compression}
\author{
    Yuyang Xu\textsuperscript{1,3,5,8,}\equalcontrib, 
    Yi Cheng\textsuperscript{2,}\equalcontrib,
    Haochao Ying\textsuperscript{1,4,5,}\thanks{Corresponding authors: Haochao Ying and Jian Wu.}, 
    Zhuoyun Du\textsuperscript{6,7,8}, \\
    Renjun Hu\textsuperscript{9}, 
    Xing Shi\textsuperscript{2}, 
    Wei Lin\textsuperscript{2},
    Jian Wu\textsuperscript{1,4,5,8,}\footnotemark[2]
}
\affiliations{
\textsuperscript{1}State Key Laboratory of Transvascular Implantation Devices, \\
The Second Affiliated Hospital Zhejiang University School of Medicine
\\
\textsuperscript{2}Alibaba Cloud Computing \\
\textsuperscript{3}College of Computer Science and Technology, Zhejiang University
\\
\textsuperscript{4}School of Public Health, Zhejiang University
\\
 \textsuperscript{5}Transvascular Implantation Devices Research Institute
\\
\textsuperscript{6}State Key Lab of CAD \& CG, Zhejiang University \\
\textsuperscript{7}Zhejiang Polytechnic Institute, Polytechnic Institute, Zhejiang University \\
 \textsuperscript{8}Zhejiang Key Laboratory of Medical Imaging Artificial Intelligence \\
\textsuperscript{9}School of Data Science and Engineering, East China Normal University \\
\texttt{xuyuyang@zju.edu.cn} \quad 
\texttt{haochaoying@zju.edu.cn}
}

\usepackage{bibentry}

\begin{document}

\maketitle

\begin{abstract}
Test-time scaling has proven effective in further enhancing the performance of pretrained Large Language Models (LLMs). However, mainstream post-training methods (i.e., reinforcement learning (RL) with chain-of-thought (CoT) reasoning) often incur substantial computational overhead due to auxiliary models and overthinking. In this paper, we empirically reveal that the incorrect answers partially stem from verbose reasoning processes lacking correct self-fix, where errors accumulate across multiple reasoning steps. 
To this end, we propose \ML (\M), a pluggable RL process supervision framework that enables fine-grained optimization of each reasoning step. Specifically, \M requires neither auxiliary models nor stepwise manual annotations. Instead, it leverages step-wise preference signals generated by the model itself to guide the optimization process for reasoning compression.
Experiments demonstrate that the generated reasoning sequences from \M are both accurate and succinct, effectively mitigating overthinking behaviors without compromising model performance across diverse domains and languages.
\end{abstract}

\section{Introduction}\label{sec:intro}


As Large Language Models (LLMs) and their training methodologies reach new levels of sophistication, powerful foundational models have been witnessed across different fields. 
Although LLMs exhibit impressive generalization across a wide range of tasks, their success heavily relies on pre-training with massive datasets and scaling to billions of parameters.
Consequently, recent researches~\cite{zhang2025survey, lightman2023let, snell2024scaling} have shifted priorities toward optimizing model performance at inference time, rather than exclusively emphasizing pre-training.
Test-time scaling—a strategy that 
dynamically allocates computational resources during inference—has opened up a new era for post-training optimization.

Integrating reinforcement learning (RL) with chain-of-thought (CoT) reasoning has emerged as a promising strategy to enable effective test-time scaling. In contrast to Proximal Policy Optimization (PPO)~\cite{schulman2017proximal} demands substantial computational resources, Group Relative Policy Optimization (GRPO)~\cite{shao2024deepseekmath}—a representative RL framework—combines CoT reasoning with rule-based reward signals to supervise LLMs. This approach enables autonomous exploration of reasoning pathways to align human values and even facilitates ``Aha Moments'', where the model refines its own logic. By replacing reward design from exhaustive training of the reward model with rule-based group relative advantage evaluation, GRPO achieves superior performance on complex tasks while maintaining cost efficiency. 

However, though it is widely believed that the longer the LLM reasons, the more accurate it performs, GRPO and similar RL-based methods face a critical limitation~\cite{kumar2025overthink}: \textit{the tendency to overthink}, even when solving simple tasks. We hypothesize that rule-based reward approaches (\textit{e.g.,} GRPO)—which rely solely on sparse reward signals indicating final answer correctness—fail to penalize suboptimal intermediate reasoning pathways. This limitation creates a permissive exploration environment where policy networks are incentivized to generate excessively elaborate reasoning chains, ultimately manifesting as overthinking. Such behavior not only undermines efficiency during inference but can also reduce accuracy by introducing and accumulating errors in redundant reasoning.

On the other hand, process supervision has proven to be an effective post-training methodology for improving LLM reasoning capabilities at inference time. Prominent techniques include Monte Carlo Tree Search (MCTS) and Process Reward Models (PRMs) for strategic exploration of reasoning pathways by evaluating intermediate reasoning steps. However, these methods typically require computationally intensive multi-rollout generation or the deployment of external open-source/closed-source models as reasoning evaluators~\cite{lightman2023let, yun2025med, lai2024step, kumar2024training}, which obscures the elegance of GRPO's reward signal design. This raises a critical question: whether we can develop a framework that simultaneously achieves two desiderata—(1) \textit{simplicity in reward formulation}, and (2) \textit{step-wise granular dense supervision}.

To this end, we draw inspiration from recent studies suggesting that well-pretrained LLMs inherently possess endogenous reward models that unify diverse post-training methodologies~\cite{rafailov2023direct, li2025generalist, wang2025implicit}. 
Building on this insight, we propose step-wise Verbal Value Probing (VVP), which leverages the policy model itself to evaluate reasoning process values. Instead of linear probes, VVP concatenates context-aware answer templates to rollout responses, prompting the LLM to self-assess the current reasoning state through its native output distribution. Our preliminary findings demonstrate that CoT reasoning exhibits pathological toxicity in specific query domains, where extended reasoning trajectories paradoxically degrade solution quality and trigger overthinking behaviors.

Building upon the VVP method, we revisit the Generalized Advantage Estimator (GAE) and introduce a pluggable fine-grained dense supervision signal for CoT-based reinforcement learning (RL) reasoning. We propose \M, \ML, which do not require either auxiliary deep learning-based models or human annotations for step-wise supervision. By integrating rule-based fine-grained supervision through a computationally efficient formulation, our approach constrains reasoning trajectories to maintain coherence while mitigating overthinking. This is achieved by dynamically calibrating the advantage function with VVP-derived state evaluations, ensuring gradient updates prioritize logically consistent reasoning steps over redundant explorations. To validate the efficacy of our proposed pluggable step-wise supervision method \M, we conduct comparative experiments across multi-task, multilingual benchmarks. Experimental results demonstrate that \M achieves dual advantages: Preservation/enhancement of reasoning performance and reasoning compression. Our main contributions are as follows:

\begin{itemize}
    \item We propose a model-free reasoning process value-estimation method, named VVP, which can provide a solution for estimating step-wise in reasoning.
    \item We propose the \M framework based on VVP for step-wise dense supervision and reasoning compression, eliminating the need for both auxiliary models and step-wise human annotation.
    \item Experiments across different tasks and languages prove that \M can mitigate the overthinking problem while maintaining the problem-solving capability.
\end{itemize}

\section{Related Work}

\subsection{Reinforcement Learning for Value Alignment}

After pre-training, Supervised Fine-Tuning (SFT) endows the model with fundamental instruction-following capabilities through guided instruction, whereas Reinforcement Learning (RL) enables the model to autonomously explore the solution space while receiving supervisory signals that align its behavior with human value preferences. Building upon Markov decision processes (MDPs) and Bellman equations, the Proximal Policy Optimization (PPO)~\cite{schulman2017proximal} algorithm with Generalized Advantage Estimation (GAE)~\cite{schulman2015high} formulates step-wise rewards as the combination of immediate rewards from the reward model and discounted cumulative future returns from the critic model. 
Following PPO, new RL methods~\cite{rafailov2023direct, hu2025reinforce++, shao2024deepseekmath, yu2025dapo} are designed to reduce computational costs based on the intrinsic reward model in the policy model and rule-based reward estimation. Among them, Direct Preference Optimization~(DPO)~\cite{rafailov2023direct} simplifies the preference alignment objective through mathematical derivations, enabling SFT-like training of aligned models with minimal overhead. On the other hand, GRPO~\cite{shao2024deepseekmath} takes the overall generated sequence as a single step and replaces the reward model with rule-based Ground Truth checking, pioneering a new paradigm of RL algorithms by introducing rule-derived efficient rewards.
However, in contrast to PPO, GRPO-style methods exhibit critical limitations by delivering excessively sparse reward signals, which are constrained to assign equal reward to all tokens and fail to provide fine-grained supervision for sequence generation.

Under the guidance of DPO, an increasing number of studies have attempted to adopt the approach of utilizing signals generated by the actor model itself as reward signals for training the actor model. \citet{wang2025implicit} and \citet{li2025generalist} also demonstrated through inverse reinforcement learning that LLMs inherently contain an endogenous reward model within after the next-word prediction pre-training, which can be leveraged to optimize policy models for improved performance. However, existing approaches still predominantly rely on fine-tuning the evaluation capabilities of actor models~\cite{yuan2024self, munkhbat2025self, zhang2025process} or invoking API-based LLMs~\cite{zhang2025critique} to generate step-wise evaluations, which are subsequently utilized for iterative optimization attempts. These methodologies inevitably demand substantial computational resources due to the continuous requirement for model inference and evaluation cycles.

\begin{figure*}[t]
\centering
  \includegraphics[width=\textwidth]{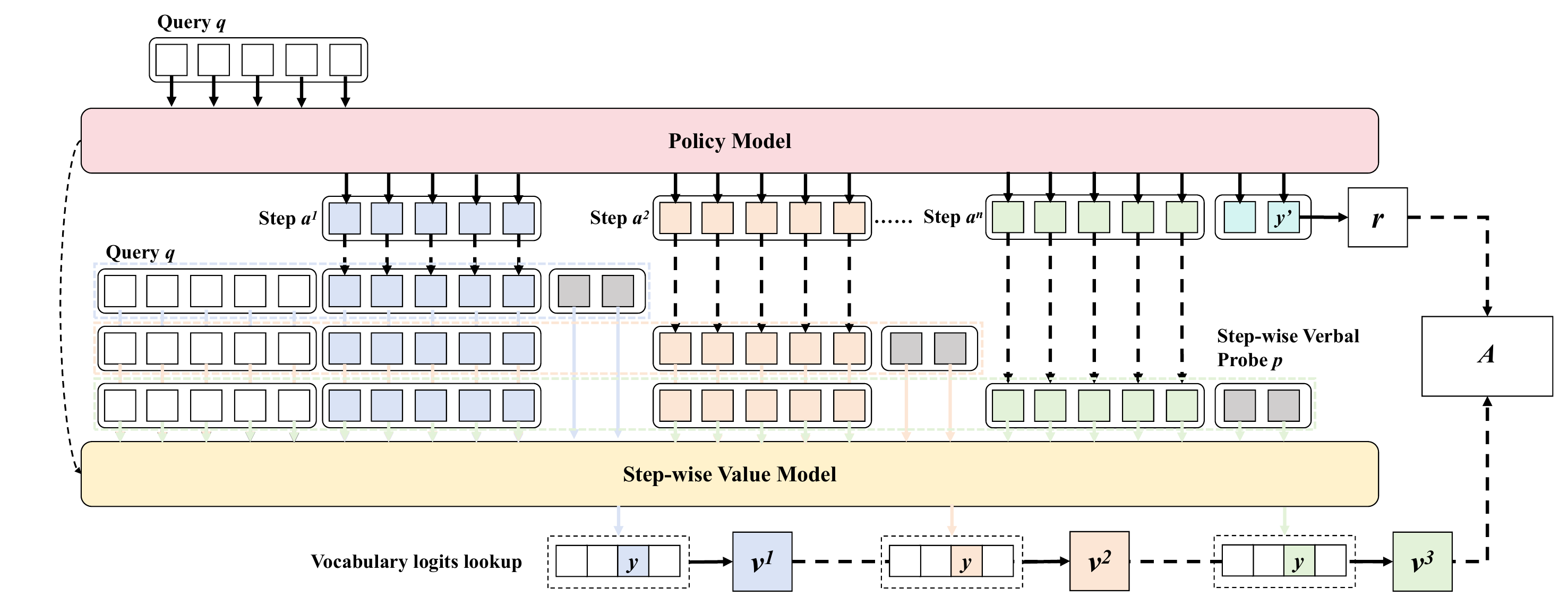} 
  \caption{Illustration of the overall pipeline of \M. For each rollout of a query example, we separate each reasoning step and utilize the step-wise Verbal Value Probing (VVP) to estimate the step-wise values.  We revisit GAE~\cite{schulman2015high} and iteratively estimate the step-wise relative advantage for computing the final loss.}
  \label{fig:main}
\end{figure*}

\subsection{Process Supervision for Reasoning Optimization}

With the proposal of chain-of-thought~\cite{wei2022chain}, LLMs expose the reasoning ability in solving complex questions. 
\citet{lightman2023let} and \citet{snell2024scaling} demonstrated that computational resources allocated to post-training/inference phases yield greater improvements in reasoning capabilities, revealing the existence of a test/inference scaling law. As a prevalent test-time scaling approach, Monte Carlo Tree Search (MCTS) methods~\cite{guan2025rstar} constructs search trees through iterative exploration-exploitation strategies, prioritizing critical node expansions based on probabilistic evaluations.
Revisiting the LLM-as-a-judge paradigm~\cite{zheng2023judging}, process reward model (PRM)~\cite{lightman2023let, yun2025med, wang2025visualprm} constitutes an intuitive approach for process evaluation for MCTS through auxiliary training of a step-wise evaluation module that performs binary classification on reasoning steps. Though PRM can provide dense supervision for reasoning rollouts, the training phase of PRM requires labor-intensive annotation for each step. Several methods~\cite{lai2024step} also apply powerful API-based LLMs as PRM and generate step-wise preference labels for the attempting rollouts of the detected error steps. Still, they require auxiliary computational resources during inference.

\subsection{Overthinking in RL-based reasoning models}

Reasoning methods akin to GRPO have endowed LLMs with the capability to experience insightful moments for self-correcting reasoning errors. Despite this advancement, flawed reasoning processes may still generate redundant inference steps, and more critically, compromise the final outcomes. \citet{su2025between} employ a SimPO-based training framework with constrained output length to reveal the existence of redundant reasoning patterns of LLMs. \citet{kumar2025overthink} introduced an adversarial approach through ``bait questions'' that deliberately induce excessive, unnecessary token generation in CoT sequences, amplifying computational overhead, latency, and operational costs. There are mainly two types of methods in order to solve the overthinking problem. Methods like \citet{yu2025dapo} and \citet{xiao2025fast} define a hard sequence length in the loss function, while methods like \citet{yang2502towards} collect typical ``seed data'' for different tasks accordingly to constrain the sequence length. We identify a critical limitation in current approaches to mitigating the overthinking issue in large language models (LLMs). Specifically, existing methods predominantly rely on heuristic thresholds for generation sequence lengths, which necessitates extensive manual tuning and fails to harness the LLM's intrinsic capability for adaptive learning of optimal reasoning depth from data. 

Therefore, there exists a pressing need to develop an efficient and resource-conscious methodology that addresses the overthinking phenomenon in LLMs arising from RL paradigms. This approach should ideally maintain the reward evaluation simplicity characteristic of rule-based methods like GRPO while simultaneously enabling effective process supervision mechanisms.

\section{Methodology}

\begin{figure*}[t]
    \centering
    \begin{subfigure}[b]{0.33\textwidth} 
        \includegraphics[width=\textwidth]{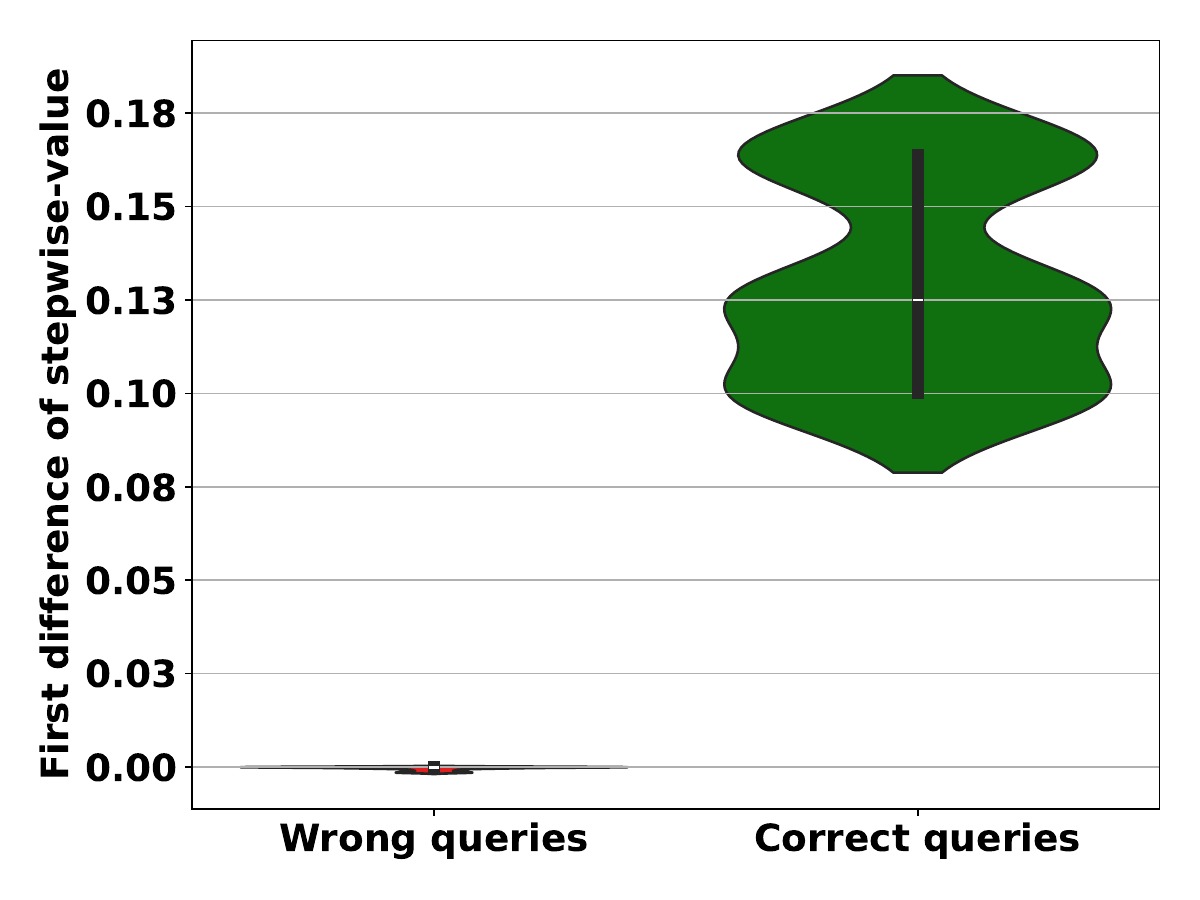} 
        \caption{Distribution of DAPO dataset.}
    \end{subfigure}
    \hfill 
    \begin{subfigure}[b]{0.33\textwidth} 
        \includegraphics[width=\textwidth]{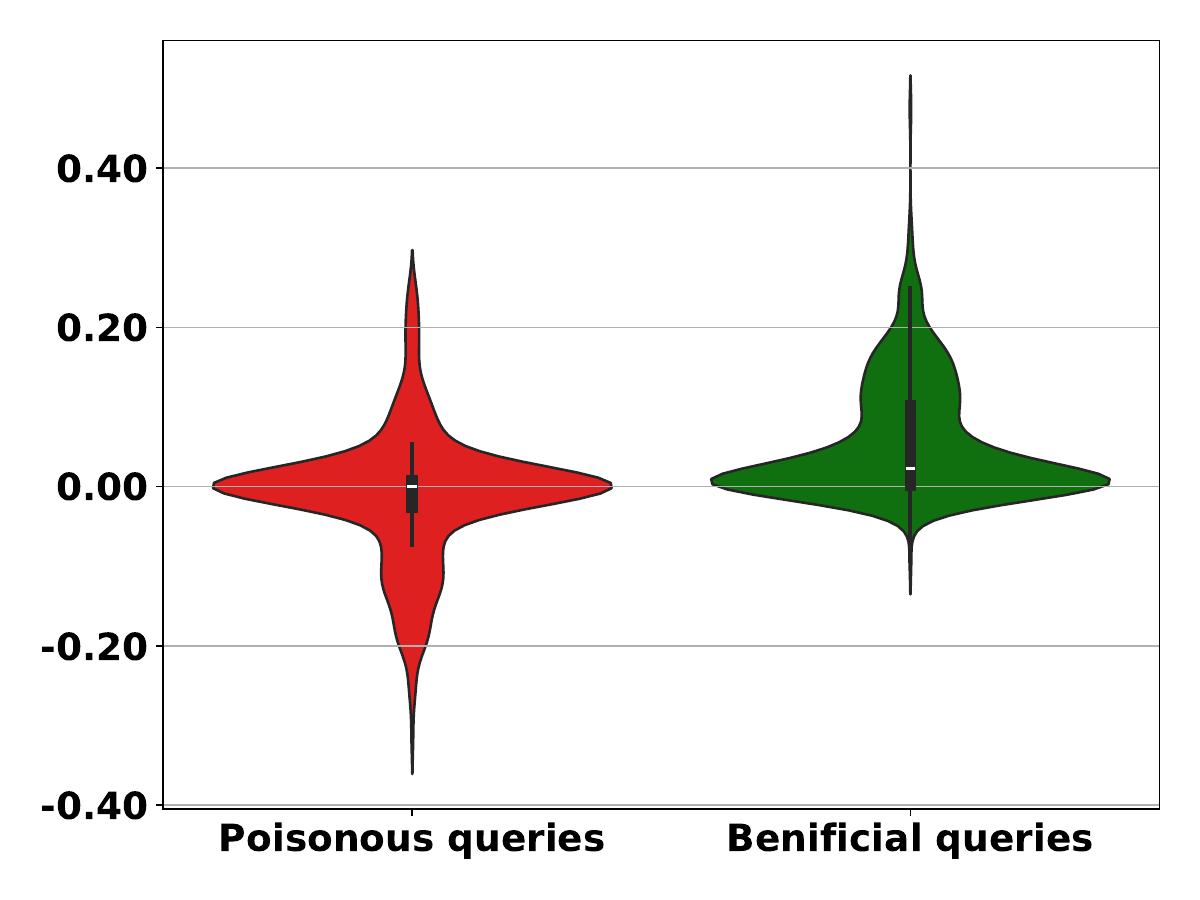} 
        \caption{Distribution of MedQA-en dataset.}
    \end{subfigure}
    \hfill 
    \begin{subfigure}[b]{0.33\textwidth} 
        \includegraphics[width=\textwidth]{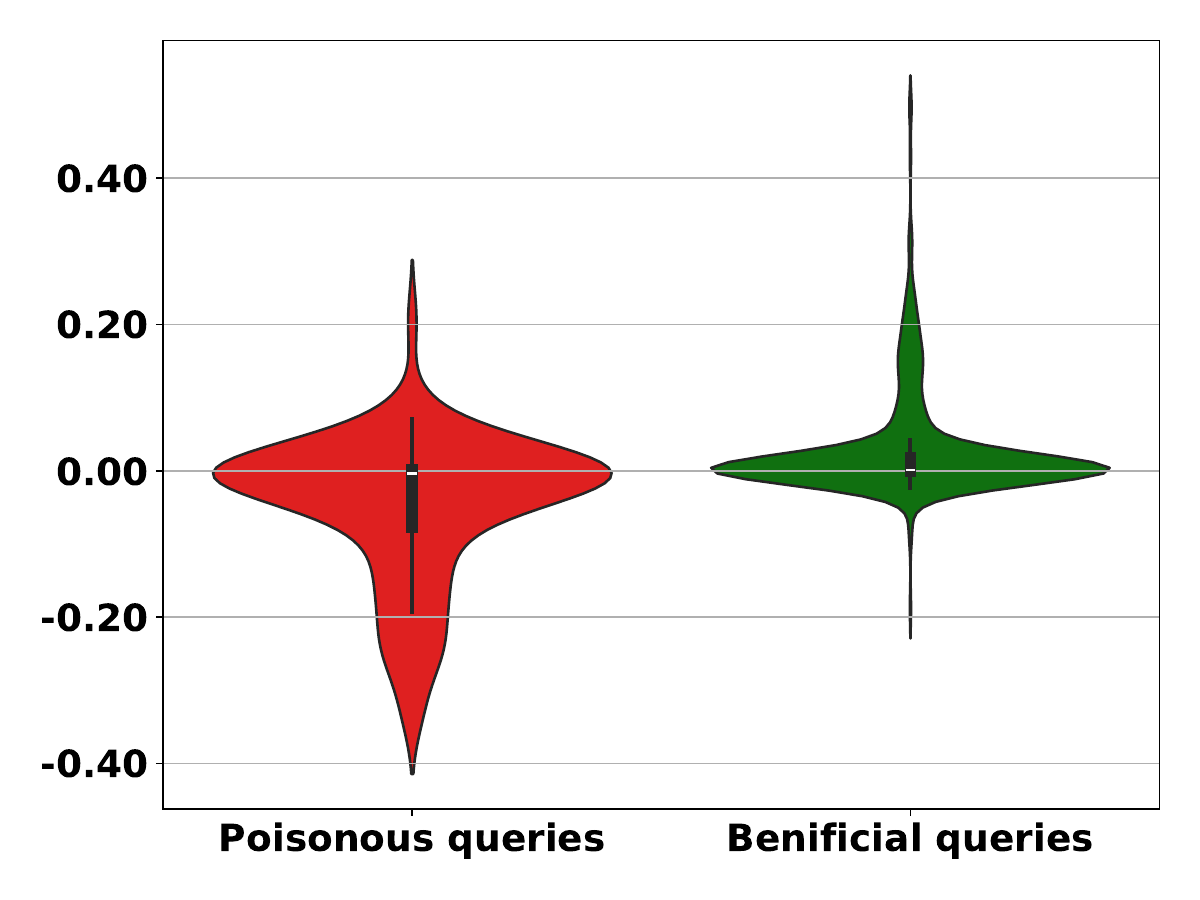} 
        \caption{Distribution of MedQA-zh dataset.}
    \end{subfigure}
    \caption{
    The distribution of the averaged first central difference of step-wise values over the response sequences. Specifically, we instruct the LLM to answer the question in both CoT and plain ways. We define queries that can be correctly answered through direct inference but fail to retrieve accurate solutions when using CoT prompting as CoT-poisonous (overthinking) queries. In contrast, we define queries that cannot be accurately answered through direct inference yet yield accurate solutions when employing CoT reasoning as CoT-beneficial (baseline) queries. Finally, we estimate the step-wise values according to the Preference Verbal Probing in \ref{sec:method_verbal} and visualize the distribution of the first central difference in the sequence. As there are no CoT poisonous samples in the DAPO dataset, we take the queries that can be correctly/wrongly answered as an example.
    }\label{fig:sv_dist}
\end{figure*}

In order to solve the overthinking problem with dense upervision at a low computation cost, we propose \M, \ML. In this section, we elaborate on the details of \M. First, we give the definition of the notations used. Then we introduce the step-wise preference estimation method, Verbal Value Probing (VVP), and conduct a pilot experiment in order to explore the pattern of step-wise values with overthinking queries. Finally, we illustrate the overall pipeline of the proposed \M, which is shown in Fig.~\ref{fig:main}. 

\subsection{Preliminaries}\label{sec:method_prelim}

In this subsection, we provide the definition of notations used in the following contents.

We define the input query token IDs to the LLM as $q \in \mathbb{N}^{l_q}$, where $l_q$ is the sequence length of the input query. 
We denote the output response IDs with a length of $l_r$ as $o = [s_1, s_2, ..., s_T, s_{c}, y] \in \mathbb{N}^{l_r}$. Here $s_n \in \mathbb{N}^{l_n} (n\in[1,T])$ refers to the ID sequence of the $n$-th reasoning step with the length of $l_n$ in response $o$. $s_c$ stands for the drawn conclusion format (\eg ``\textit{Thus, the answer is \{$\cdot$\}}''), and $y$ refers to the ground-truth (GT) string that can be validated with the rule-based methods.

For the classical PPO~\cite{schulman2017proximal} method GAE~\cite{schulman2015high}, each policy update iteration evaluates action values by combining the immediate reward obtained from the current action with the cumulative discounted future rewards. For LLMs, we define the step $t$ hidden state $h_t$ as the overall past context, that is $h_t = [p, s_1, s_2, ..., s_t]$, and the action taken at step $t$ as $s_{t+1}$. The optimization goal of the trained policy LLM $\pi_\theta$ can be formulated as:
\begin{equation}
    \mathcal{L}_{GAE} = \mathbb{E}_t[log\pi_\theta(s_{t+1}|h_t)\cdot \sum_{i=0}^{T-t}(\gamma \lambda)^i(r_t+\gamma v_{t+1} - v_t)],
\end{equation}
\noindent where $r_t$ and $v_t$ are the instant reward and the future profit when action $s_{t+1}$ is taken, respectively. $\gamma \in [0,1]$ is the discount factor and $\lambda \in [0,1]$ is the mixing factor of GAE, which controls the bias-variance tradeoff.

\subsection{Step-wise Preference Estimation}\label{sec:method_verbal}


We posit that the phenomenon of overthinking in rule-based RL methods like GRPO stems from their exclusive focus on the final correctness of entire sequences, which results in sparse binary (0/1) supervisory signals applied uniformly across all tokens. This architectural limitation inherently compromises the ability to enforce intermediate reasoning constraints, thereby enabling the policy model to engage in excessive exploration during the sequential decision-making process. On the other hand, extensive computation and labor annotation resources are required for the application of PRM~\cite{lightman2023let}. Thereby, a paradigm shift is required towards an approach that ideally maintains the reward evaluation simplicity characteristic of rule-based methods, while simultaneously enabling effective process supervision mechanisms.

Concurrently, DPO~\cite{rafailov2023direct}, \citet{li2025generalist} and \citet{wang2025implicit} theoretically and practically demonstrate through inverse reinforcement learning that a powerful, generalist reward model is inherently latent within any LLM trained via standard next-token prediction pre-training. Following this, we propose step-wise Verbal Value Probing (VPP) for step-wise preference estimation:
\begin{equation}
\begin{aligned}
    u_t =& \pi_\theta(g|h_t, s_c) \\
        =& \pi_\theta(g|q, s1, s2, ..., s_t, s_c),
\end{aligned}
\end{equation}
\noindent where $u_t$ is the evaluated preference at $t$-th step. Typically, PRM is trained to inspect the probability ``+'' and ``-'' at the next token of each reasoning step. Following this, we propose to use in-context learning by adding the drawn conclusion format $s_c$ at the end of each step, in order to instruct the LLM to estimate the future value at the $t$-th step at the very next token by itself. Given a well-pretrained LLM, the inspected probability of VVP indicates the GT retrieval probability merely with steps before the $t$-th step. Thus, we utilize the conclusion format $s_c$ as the probe to elicit the pretrained LLM output of the transitional hidden-state of GT. In this way, we take the inspected probability as the evaluated human coarse preference at step $t$ according to GT $y$.

In order to investigate the mutual relation between the step-wise preference value and overthinking, we first instruct the LLM to answer the questions in direct inference style and CoT reasoning style. We define CoT-poisonous queries as inputs for which direct inference yields the correct answer, but CoT reasoning does not. In contrast, CoT-beneficial queries are those that are only correctly answered when CoT reasoning is applied. Typically, we believe that the CoT reasoning from poisonous queries are a representation of overthinking while the beneficial queries act as a positive baseline. Thus, we then apply the proposed VVP to estimate the step-wise values for each rollout sequence. Here, we compute the first difference of each step as a metric to represent the tendency of the step-wise values over the reasoning sequence. For MedQA datasets, we obtain 452 and 505 queries for the MedQA-en dataset and 178 and 868 queries for the MedQA-zh dataset. We plot the distribution violin figure of the three datasets. As there is no poisonous queries detected in the DAPO dataset, we use the queries that can and cannot be answered correctly in CoT reasoning style instead. As shown in Fig.~\ref{fig:sv_dist}, we observe that when CoT reasoning leads to errors in tasks that would otherwise be correctly solved by LLMs, our proposed VVP manifests this phenomenon as a fluctuating or decremental trend in the step-wise value estimates throughout the reasoning trajectory. This observation provides empirical validation for our hypothesis that overthinking stems from error-prone reasoning steps, which can be caused by sparse supervisory signals, while this issue can be mitigated through dense step-wise supervision that explicitly penalizes suboptimal reasoning pathways.

\subsection{Self-traced Advantage Computation}

PPO~\cite{schulman2017proximal,schulman2015high} achieves dense supervision across sequential decision-making through iterative estimation of action-state immediate rewards and cumulative future returns via dedicated reward and critic networks. However, this auxiliary model architecture incurs substantial computational overhead due to the necessity of reward and value estimation for the advantage computation over all of the possible states and actions. In the Bellman function, future benefits are estimated by the value of the $t$-th step $v_t$. For the step-wise preference, $u_t$ evaluates the probability of retrieving the GT $y$, which we believe can be regarded as the future benefits. Since 
$u_t$ can be efficiently retrieved via step-wise prefill processes and rule-based matching, VVP emerges as a computationally efficient yet effective solution for estimating the step-wise value and implementing dense supervision in LLM post-training. Thus, we reformulate the advantage computation formulation as:
\begin{equation}\label{eq:gae}
    \begin{aligned}
        A_{SSPO}^{j}(t) =& \sum_{i=0}^{T-t}(\gamma\lambda)^{i}\delta_{t+i}^{j} \\
        =& \sum_{i=0}^{T-t}(\gamma\lambda)^{i}(\hat{r}^j + \gamma\hat{v}^{j}_{t+1} - \hat{v}^{j}_{t}),
    \end{aligned}
\end{equation}
\begin{equation}\label{eq:group_norm}
    \hat{r}^j = \frac{(r^j - \bar{r})}{\sigma(\{r^j, j \in [1,n]\})},
\end{equation}
\begin{equation} \label{eq:value_norm}
    \hat{v}^{o}_{t} = \frac{(u^o_t - \bar{r})/}{\sigma(\{r^j, j \in [1,n]\})},
\end{equation}
\noindent where $\hat{r}^j$ stands for the normalized GRPO reward $r^j \in \{0, 1\}$ of the $j$-th rollout among total $n$ rollouts with an average of $\bar{r}$ and standard deviation $\sigma(r)$. 

Specially, for PPO, the optimization target for the critic model is to minimize the difference between $\mathbb{E}(r_t +v_{t+1})$ and $v_t$ in order to guarantee the convergence of Bellman function. For the last step $T$, we suppose $v_{T+1} = 0$, which means $v_{T} = \mathbb{E}(r_t)$. Thereby, we formulate the reward signal for all rollout trajectories as a binary indicator (0/1) derived from rule-based answer matching. Intuitively, when applying our proposed VVP methodology, under the theoretical foundation of the Law of Large Numbers, the expectation of this binary signal reward demonstrates convergence to $u_T$, the probability of generating GT tokens $y$ as the number of rollout samplings approaches infinity. 

In order to compute the group relative advantage, the reward $r^j$ is normalized over the rollout batch with a batch size of $n$ rollouts, as shown in Eq.~\ref{eq:group_norm}. In order to guarantee the convergence of the Bellman euqation $v_t = \mathbb{E}(r_t +v_{t+1})$, we also normalize and reformulate the value, as shown in Eq.~\ref{eq:value_norm}.

\subsection{Error Step Pruning}

The core principle of the Markov assumption suggests, future states depend solely on the current state, independent of historical states. In the context of LLMs, the current state encompasses the input query $q$ and all previously generated reasoning steps $[s_1, s_2, ..., s_t]$. Consequently, the cumulative impact of past errors may propagate and compromise subsequent reasoning steps. Therefore, we propose the Error Step Pruning strategy. We believe that the value sequence corresponding to a preferred reasoning step sequence should exhibit a monotonically increasing trend. In this way, we implement the Error Step Pruning by first identifying the initial inflection point where the step-wise value trajectory begins to decline, followed by pruning all subsequent reasoning steps to exclude them from subsequent gradient updates, that is

\begin{equation}\label{eq:neighborcut}
    \begin{aligned}
    \mathcal{L}_{SSPO} =& -\mathbb{E}[\mathbb{I}(s_t^j)log\pi_\theta(s_{t+1}^j|h_t^j)\cdot \sum_{i=0}^{T-t}(\gamma \lambda)^{i}\tilde{\delta}_{t+1}^j], \\
    \tilde{\delta}_{t+1}^j = & \hat{r}^j + \mathbb{I}(s_t^j)(\gamma\hat{v}^{j}_{t+1} - \hat{v}^{j}_{t}), \\
    \mathbb{I}(s_t^j) =& \begin{cases}
        1 & \text{if } t \leq min(\{e|v_e \leq v_{e-1}\}), \\
        0 & \text{if } t > min(\{e|v_e \leq v_{e-1}\}),
    \end{cases}
    \end{aligned}
\end{equation}

\noindent where for those value declinging steps, the advantage function $\tilde{\delta}_{t+1}^j$ degrade into the vanilla GRPO.

\section{Experiments}

\begin{table}
\centering
\begin{tblr}{
  cells = {c},
  cell{1}{1} = {c=2}{},
  cell{2}{1} = {c=2}{},
  cell{3}{1} = {r=3}{},
  cell{6}{1} = {r=3}{},
  vline{2,4} = {1-8}{},
  vline{2-4} = {3,8}{},
  vline{3-4} = {1-8}{},
  hline{1-3,6,9} = {-}{},
}
\textbf{Benchmark } &                 & \textbf{Train} & \textbf{Test} \\
\textbf{DAPO}       &                 & 17917          & 960           \\
\textbf{MedQA-en}   & \textbf{U.S.}   & 10178          & 1273          \\
                    & \textbf{T.W.C.} & 11298          & 1413          \\
                    & \textbf{All}    & 21476          & 2686          \\
\textbf{MedQA-zh}   & \textbf{M.L.C.} & 27400          & 3426          \\
                    & \textbf{T.W.}   & 11298          & 1413          \\
                    & \textbf{All}    & 38698          & 4839          
\end{tblr}
\caption{Statistics of the used benchmarks.}\label{tab:sta}
\end{table}

\begin{table*}
\centering
\begin{tblr}{
  cells = {c},
  cell{1}{1} = {r=2}{},
  cell{1}{2} = {c=2}{},
  cell{1}{4} = {c=2}{},
  cell{1}{6} = {c=2}{},
  vline{2-6} = {1}{},
  vline{2,4,6} = {2}{},
  vline{2,4,6} = {3-6}{},
  hline{1,3,5,7} = {-}{},
}
\textbf{Benchmark}           & \textbf{AIME24}  &                     & \textbf{MedQA-en} &                              & \textbf{MedQA-zh} &                              \\
                             & \textbf{ACC $\uparrow$}     & \textbf{Resp. Len. $\downarrow$} & \textbf{ACC $\uparrow$}      & \textbf{\textbf{Resp. Len. $\downarrow$}} & \textbf{ACC $\uparrow$}      & \textbf{\textbf{Resp. Len. $\downarrow$}} \\
GRPO                         & 18.12\%          & 1439.07             & \textbf{71.10\%}  & 425.49                       & \textbf{86.65\%}  & 298.73                       \\
\textbf{\textbf{SSPO(GRPO)}} & 1\textbf{9.58\%} & \textbf{907.84}     & 70.40\%           & \textbf{410.29}              & 86.34\%           & \textbf{292.50}              \\
DAPO                         & 14.37\%          & 1035.94    & \textbf{72.04\%}  & 406.82                       & 86.80\%           & \textbf{303.01}              \\
\textbf{\textbf{SSPO(DAPO)}} & \textbf{14.68\%} & \textbf{999.18}             & 71.63\%           & \textbf{393.58}              & \textbf{87.19\%}  & 333.41                       
\end{tblr}
\caption{Comparison result of SSPO and naive GRPO/DAPO on Accuracy (ACC) and averaged Response Length (Resp. Len.). The best performances are marked in bold.}\label{tab:result}
\end{table*}

\begin{figure*}[t]
    \centering
    \begin{subfigure}[b]{0.33\textwidth} 
        \includegraphics[width=\textwidth]{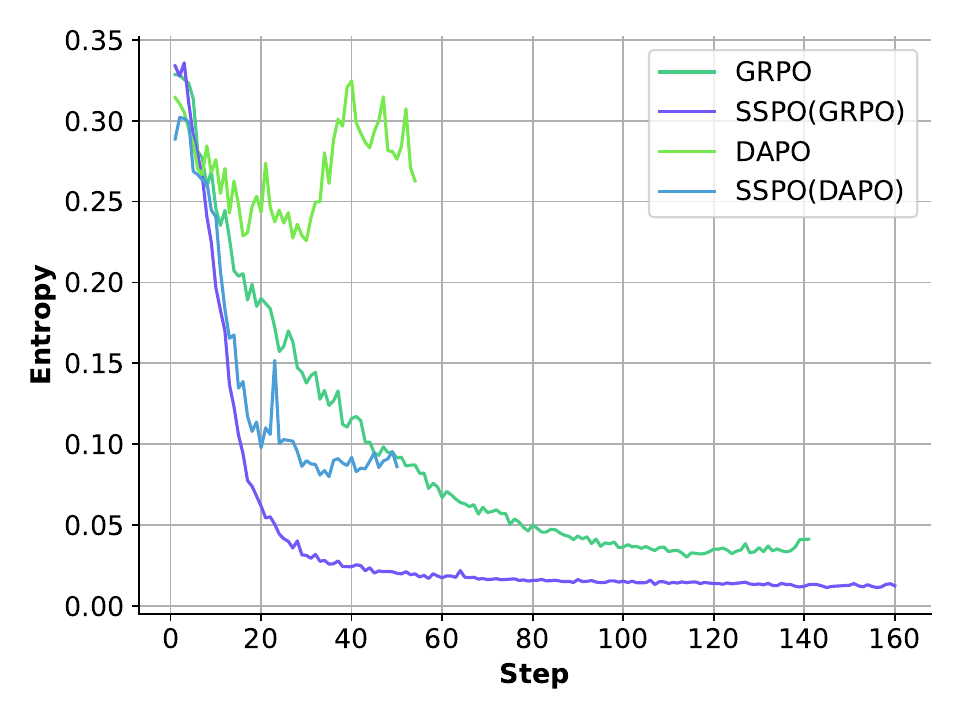} 
        \caption{Entropy Trajectory of DAPO dataset.}
    \end{subfigure}
    \hfill 
    \begin{subfigure}[b]{0.33\textwidth} 
        \includegraphics[width=\textwidth]{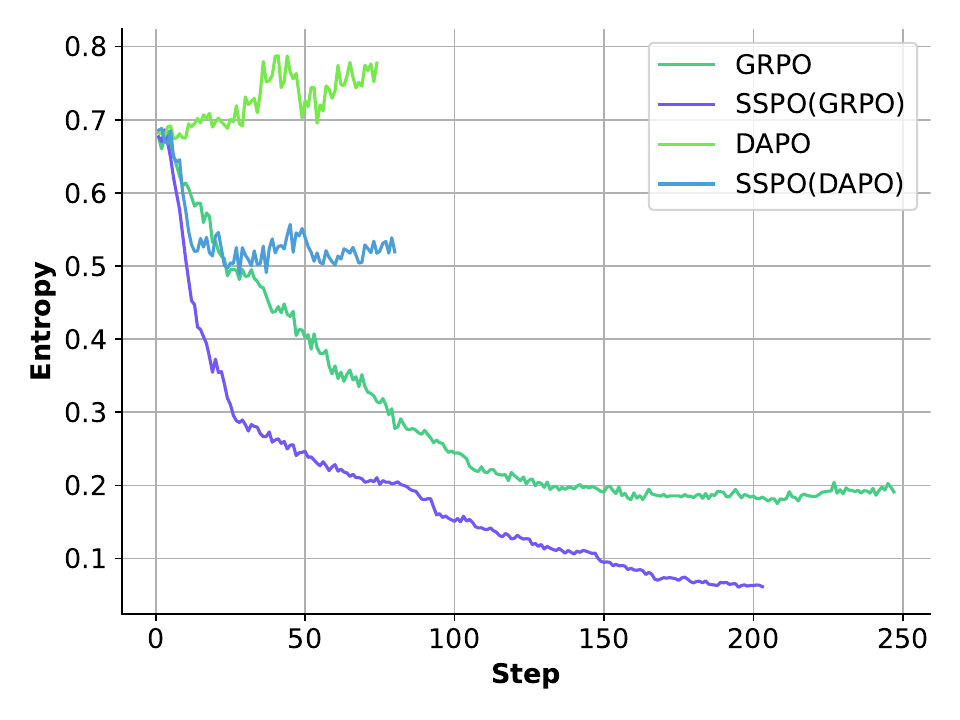} 
        \caption{Entropy Trajectories of MedQA-en dataset.}
    \end{subfigure}
    \hfill 
    \begin{subfigure}[b]{0.33\textwidth} 
        \includegraphics[width=\textwidth]{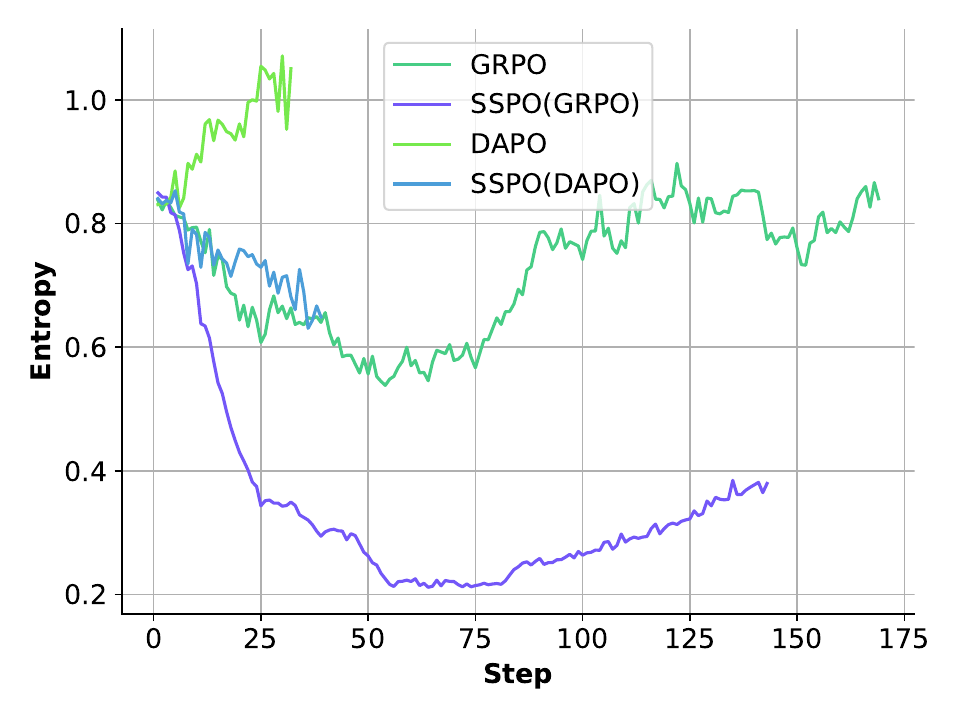} 
        \caption{Entropy Trajectory of MedQA-zh dataset.}
    \end{subfigure}
    \caption{
    The Entropy Trajectory over the training steps of the three datasets. We can see that the entropy trajectories of \M remain below the corresponding baseline, which means that the provided dense signal
    }\label{fig:entro_track}
\end{figure*}

\begin{figure*}[t]
\centering
  \includegraphics[width=0.85\textwidth]{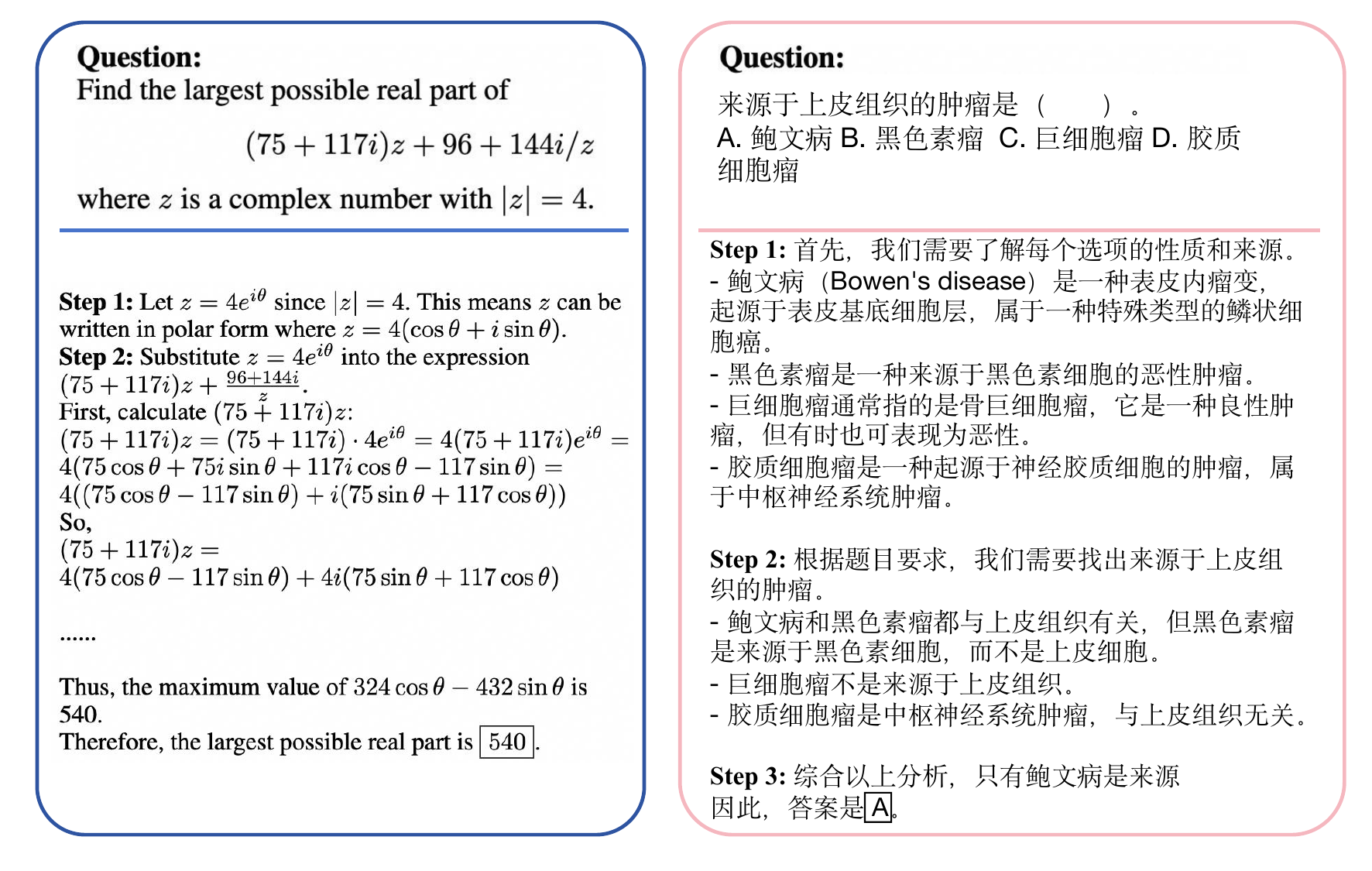} 
  \caption{Examples of reasoning process from DAPO dataset and MedQA-zh dataset.}
  \label{fig:case}
\end{figure*}

In this section, we first introduce the experimental setup including the tested LLM model and benchmarks. Then, we present the main evaluation outcomes by comparing the \M with a vanilla rule-based RL method across different domains and languages. Moreover, we analyze the entropy trajectory to investigate the step-wise supervision of \M. Finally, case study is also conducted to present the reasoning process after \M.

\subsection{Experiment Setup}\label{sec:exp_setup}

\noindent \textbf{Models and Implementations}. 
Owing to limitations in computational resources, the Qwen2.5-7B-Instruct~\cite{qwen2.5} model is predominantly employed as the foundational architecture for training the actor/policy model. We mainly adopt the rule-based RL methods, GRPO~\cite{shao2024deepseekmath} and DAPO~\cite{yu2025dapo}. Specially, DAPO explicitly incorporates a response-length regularization term in the loss function. In order to demonstrate the inherent capacity of \M without confounding factors, we systematically removed this length-based regularization component in the \M(DAPO) experimental configuration. This methodological distinction enables direct comparison of the response-length regularization and \M.
For all the conducted experiments, we adopt VeRL~\cite{sheng2025hybridflow} as the training framework. For each example, we generate 16 different responses for group relative reward computation. We set the learning rate as $5e-7$ with the global batch size as 512.

\noindent \textbf{Benchmarks}. In order to evaluate the proposed \M, we train the LLM on Mathematical and Medical tasks. 

For the Mathematical reasoning task, we adopt the DAPO dataset from~\citet{yu2025dapo}. We utilize 1,791,700 mathematical QA problems in the DAPO-Math-17k dataset as the training corpus. As each of the individule QA is repeated 100 times in the dataset, we filter out the repeated QA samples and use the deduplicated 17917 QAs as the final training set. For the test set, we adopt 960 mathematical QAs in the AIME-2024 dataset.

For Medical reasoning task, we apply the MedQA~\cite{jin2021disease} as the training and testing corpus. MedQA is a medical multiple-choice question-answering dataset in the format of multiple-choice questions. The questions are collected from medical licensing examinations in the United States (U.S.), Mainland China (M.L.C.), and Taiwan China (T.W.C.). These exams are designed to assess the professional knowledge and clinical decision-making abilities of medical practitioners. We combine the english version of exams from Taiwan and exams from US as the MedQA-en dataset, and the chinese version of Taiwan exams and the exams from Mainland China as MedQA-zh. We utilize the train part as the training set and utilize the dev and the test part as the testing set. The statistics of the three datasets are shown in Tab.~\ref{tab:sta}.

\noindent \textbf{Evaluation}. As \M is a pluggable method for any rule-based Policy Optimization problem, we adopt GRPO~\cite{shao2024deepseekmath} and DAPO~\cite{yu2025dapo} as baseline methods and compare the GRPO-based and DAPO-based \M for evaluation. For both the mathematical and the medical task, we consider a response to be correct if it exactly matches the GT; all other responses are classified as incorrect. We mainly evaluate the effectiveness of the baselines and \M with the accuracy as well as the averaged response length for problem-solving capability and overthinking.

\subsection{Main Results}\label{sec:exp_result}



As shown in Tab.~\ref{tab:result}, our proposed model \M remarkably compresses the reasoning process in CoT-based responses while achieves comparable or improved performance compared to the baseline methods across cross-disciplinary and multilingual datasets. To be specific, we improved the reasoning performance of GRPO on 7B LLM by 1.46\% wile compress the reasoning process by 36.91\%. Notably, the response lengths generated by our proposed \M method-trained LLMs demonstrate comparable response sequence length to those produced by DAPO with explicit response length regularization loss. This observation indicates that our framework can automatically adapt to varying reasoning length requirements across different tasks, eliminating the need for manual specification of required response lengths. 

Building upon the rule-based reward methodology exemplified by GRPO, our proposed \M framework demonstrates remarkable efficiency in generating step-wise dense supervision signals. This dense supervisory mechanism effectively mitigates erroneous and redundant reasoning steps, thereby significantly alleviating overthinking phenomena in LLMs.
Critical to our approach is the dual-advantage formulation in Eq.~\ref{eq:gae}: The term $\hat{r}^j$ provides within-sample advantage estimation across different rollouts, while $\hat{v}^{j}_{t+1} - \hat{v}^{j}_{t})$ contributes temporal advantage differentiation within individual rollouts. This dual-mechanism operates analogously to differential analysis in Sec. Step-wise preference Estimation, enabling systematic identification of potential erroneous reasoning steps through comparative assessment of value function dynamics.

\subsection{Analysis of Entropy Trajectory during training}\label{sec:exp_entro}

To further investigate the effect of the dense supervision from \M, we visualize the entropy trajectory during the training process, which are shown in Fig.~\ref{fig:entro_track}.
We observe that the entropy curves of \M exhibit the same as high as the initial values of the baseline but gradually decrease over time, consistently maintaining lower levels compared to the corresponding baseline. We attribute this behavior to the dense supervisory signals inherent in \M, which effectively constrain the model's reasoning trajectory during mid-to-late training stages. This mechanism dynamically balances exploration and exploitation by progressively aligning the model's decision-making process with human preferences, thereby preventing erroneous "aha moments" that could lead to overthinking. The entropy reduction pattern suggests that our framework successfully suppresses excessive exploration in later phases while maintaining sufficient initial diversity, ultimately achieving more stable and result-aligned reasoning pathways.

\subsection{Case Study}\label{sec:exp_case}

In this subsection, we present the case study from an English Mathematical dataset and a Chinese Medical dataset, as shown in Fig.~\ref{fig:case}. As most of the questions in MedQA concern knowledge acquisition and process, the response lengths of the two MedQA datasets are much shorter than the response lengths in DAPO. This also proves that without specific human regulation, \M can adaptively learn the necessary response length for different questions.

\section{Conclusion}


We believe that sparse supervision from rule-based RL methods like GRPO is one of the causes leading to the overthinking problem of reasoning in LLMs. Despite the heavy computational consumption of the existing process supervision methods, we propose step-wise Verbal Value Probing (VVP) as a model-free step-wise value estimation method. Based on VVP, we construct \M, \ML that offers an efficient pluggable dense process supervision solution for the rule-based RL methods.
The proposed method achieves comparable or superior performance to naive baselines while significantly reducing the response length of baseline models across different tasks and languages, offering new insights into process supervision during LLM post-training.

\bibliography{aaai2026}

\begin{thebibliography}{29}
\providecommand{\natexlab}[1]{#1}

\bibitem[{Guan et~al.(2025)Guan, Zhang, Liu, Shang, Sun, Zhu, Yang, and Yang}]{guan2025rstar}
Guan, X.; Zhang, L.~L.; Liu, Y.; Shang, N.; Sun, Y.; Zhu, Y.; Yang, F.; and Yang, M. 2025.
\newblock rStar-Math: Small LLMs Can Master Math Reasoning with Self-Evolved Deep Thinking.
\newblock \emph{arXiv preprint arXiv:2501.04519}.

\bibitem[{Hu et~al.(2025)Hu, Liu, Xu, and Shen}]{hu2025reinforce++}
Hu, J.; Liu, J.~K.; Xu, H.; and Shen, W. 2025.
\newblock Reinforce++: An efficient rlhf algorithm with robustness to both prompt and reward models.
\newblock \emph{arXiv preprint arXiv:2501.03262}.

\bibitem[{Jin et~al.(2021)Jin, Pan, Oufattole, Weng, Fang, and Szolovits}]{jin2021disease}
Jin, D.; Pan, E.; Oufattole, N.; Weng, W.-H.; Fang, H.; and Szolovits, P. 2021.
\newblock What disease does this patient have? a large-scale open domain question answering dataset from medical exams.
\newblock \emph{Applied Sciences}, 11(14): 6421.

\bibitem[{Kumar et~al.(2025)Kumar, Roh, Naseh, Karpinska, Iyyer, Houmansadr, and Bagdasarian}]{kumar2025overthink}
Kumar, A.; Roh, J.; Naseh, A.; Karpinska, M.; Iyyer, M.; Houmansadr, A.; and Bagdasarian, E. 2025.
\newblock Overthink: Slowdown attacks on reasoning llms.
\newblock \emph{arXiv preprint arXiv:2502.02542}.

\bibitem[{Kumar et~al.(2024)Kumar, Zhuang, Agarwal, Su, Co-Reyes, Singh, Baumli, Iqbal, Bishop, Roelofs et~al.}]{kumar2024training}
Kumar, A.; Zhuang, V.; Agarwal, R.; Su, Y.; Co-Reyes, J.~D.; Singh, A.; Baumli, K.; Iqbal, S.; Bishop, C.; Roelofs, R.; et~al. 2024.
\newblock Training language models to self-correct via reinforcement learning.
\newblock \emph{arXiv preprint arXiv:2409.12917}.

\bibitem[{Lai et~al.(2024)Lai, Tian, Chen, Yang, Peng, and Jia}]{lai2024step}
Lai, X.; Tian, Z.; Chen, Y.; Yang, S.; Peng, X.; and Jia, J. 2024.
\newblock Step-dpo: Step-wise preference optimization for long-chain reasoning of llms.
\newblock \emph{arXiv preprint arXiv:2406.18629}.

\bibitem[{Li et~al.(2025)Li, Xu, Yu, Zhang, Chen, Ling, Chao, Yuan, and Zhou}]{li2025generalist}
Li, Y.-C.; Xu, T.; Yu, Y.; Zhang, X.; Chen, X.-H.; Ling, Z.; Chao, N.; Yuan, L.; and Zhou, Z.-H. 2025.
\newblock Generalist Reward Models: Found Inside Large Language Models.
\newblock \emph{arXiv preprint arXiv:2506.23235}.

\bibitem[{Lightman et~al.(2023)Lightman, Kosaraju, Burda, Edwards, Baker, Lee, Leike, Schulman, Sutskever, and Cobbe}]{lightman2023let}
Lightman, H.; Kosaraju, V.; Burda, Y.; Edwards, H.; Baker, B.; Lee, T.; Leike, J.; Schulman, J.; Sutskever, I.; and Cobbe, K. 2023.
\newblock Let's verify step by step.
\newblock In \emph{The Twelfth International Conference on Learning Representations}.

\bibitem[{Munkhbat et~al.(2025)Munkhbat, Ho, Kim, Yang, Kim, and Yun}]{munkhbat2025self}
Munkhbat, T.; Ho, N.; Kim, S.~H.; Yang, Y.; Kim, Y.; and Yun, S.-Y. 2025.
\newblock Self-training elicits concise reasoning in large language models.
\newblock \emph{arXiv preprint arXiv:2502.20122}.

\bibitem[{Rafailov et~al.(2023)Rafailov, Sharma, Mitchell, Manning, Ermon, and Finn}]{rafailov2023direct}
Rafailov, R.; Sharma, A.; Mitchell, E.; Manning, C.~D.; Ermon, S.; and Finn, C. 2023.
\newblock Direct preference optimization: Your language model is secretly a reward model.
\newblock \emph{Advances in neural information processing systems}, 36: 53728--53741.

\bibitem[{Schulman et~al.(2015)Schulman, Moritz, Levine, Jordan, and Abbeel}]{schulman2015high}
Schulman, J.; Moritz, P.; Levine, S.; Jordan, M.; and Abbeel, P. 2015.
\newblock High-dimensional continuous control using generalized advantage estimation.
\newblock \emph{arXiv preprint arXiv:1506.02438}.

\bibitem[{Schulman et~al.(2017)Schulman, Wolski, Dhariwal, Radford, and Klimov}]{schulman2017proximal}
Schulman, J.; Wolski, F.; Dhariwal, P.; Radford, A.; and Klimov, O. 2017.
\newblock Proximal policy optimization algorithms.
\newblock \emph{arXiv preprint arXiv:1707.06347}.

\bibitem[{Shao et~al.(2024)Shao, Wang, Zhu, Xu, Song, Bi, Zhang, Zhang, Li, Wu et~al.}]{shao2024deepseekmath}
Shao, Z.; Wang, P.; Zhu, Q.; Xu, R.; Song, J.; Bi, X.; Zhang, H.; Zhang, M.; Li, Y.; Wu, Y.; et~al. 2024.
\newblock Deepseekmath: Pushing the limits of mathematical reasoning in open language models.
\newblock \emph{arXiv preprint arXiv:2402.03300}.

\bibitem[{Sheng et~al.(2025)Sheng, Zhang, Ye, Wu, Zhang, Zhang, Peng, Lin, and Wu}]{sheng2025hybridflow}
Sheng, G.; Zhang, C.; Ye, Z.; Wu, X.; Zhang, W.; Zhang, R.; Peng, Y.; Lin, H.; and Wu, C. 2025.
\newblock Hybridflow: A flexible and efficient rlhf framework.
\newblock In \emph{Proceedings of the Twentieth European Conference on Computer Systems}, 1279--1297.

\bibitem[{Snell et~al.(2024)Snell, Lee, Xu, and Kumar}]{snell2024scaling}
Snell, C.; Lee, J.; Xu, K.; and Kumar, A. 2024.
\newblock Scaling llm test-time compute optimally can be more effective than scaling model parameters.
\newblock \emph{arXiv preprint arXiv:2408.03314}.

\bibitem[{Su et~al.(2025)Su, Healey, Nakov, and Cardie}]{su2025between}
Su, J.; Healey, J.; Nakov, P.; and Cardie, C. 2025.
\newblock Between underthinking and overthinking: An empirical study of reasoning length and correctness in llms.
\newblock \emph{arXiv preprint arXiv:2505.00127}.

\bibitem[{Team(2024)}]{qwen2.5}
Team, Q. 2024.
\newblock Qwen2.5: A Party of Foundation Models.

\bibitem[{Wang et~al.(2025{\natexlab{a}})Wang, Cheng, Peng, Bao, Li, Guo, Li, Zeng, Zhou, and Qiu}]{wang2025implicit}
Wang, B.; Cheng, Q.; Peng, R.; Bao, R.; Li, P.; Guo, Q.; Li, L.; Zeng, Z.; Zhou, Y.; and Qiu, X. 2025{\natexlab{a}}.
\newblock Implicit Reward as the Bridge: A Unified View of SFT and DPO Connections.
\newblock \emph{arXiv preprint arXiv:2507.00018}.

\bibitem[{Wang et~al.(2025{\natexlab{b}})Wang, Gao, Chen, Chen, Zhu, Zhao, Liu, Cao, Ye, Zhu et~al.}]{wang2025visualprm}
Wang, W.; Gao, Z.; Chen, L.; Chen, Z.; Zhu, J.; Zhao, X.; Liu, Y.; Cao, Y.; Ye, S.; Zhu, X.; et~al. 2025{\natexlab{b}}.
\newblock Visualprm: An effective process reward model for multimodal reasoning.
\newblock \emph{arXiv preprint arXiv:2503.10291}.

\bibitem[{Wei et~al.(2022)Wei, Wang, Schuurmans, Bosma, Xia, Chi, Le, Zhou et~al.}]{wei2022chain}
Wei, J.; Wang, X.; Schuurmans, D.; Bosma, M.; Xia, F.; Chi, E.; Le, Q.~V.; Zhou, D.; et~al. 2022.
\newblock Chain-of-thought prompting elicits reasoning in large language models.
\newblock \emph{Advances in neural information processing systems}, 35: 24824--24837.

\bibitem[{Xiao et~al.(2025)Xiao, Gan, Dai, He, Huang, Li, Shu, Yu, Zhang, Jiang et~al.}]{xiao2025fast}
Xiao, W.; Gan, L.; Dai, W.; He, W.; Huang, Z.; Li, H.; Shu, F.; Yu, Z.; Zhang, P.; Jiang, H.; et~al. 2025.
\newblock Fast-slow thinking for large vision-language model reasoning.
\newblock \emph{arXiv preprint arXiv:2504.18458}.

\bibitem[{Yang et~al.()Yang, Ma, Lin, and Wei}]{yang2502towards}
Yang, W.; Ma, S.; Lin, Y.; and Wei, F. ????
\newblock Towards thinking-optimal scaling of test-time compute for llm reasoning, 2025.
\newblock \emph{URL https://arxiv. org/abs/2502.18080}, 10.

\bibitem[{Yu et~al.(2025)Yu, Zhang, Zhu, Yuan, Zuo, Yue, Dai, Fan, Liu, Liu et~al.}]{yu2025dapo}
Yu, Q.; Zhang, Z.; Zhu, R.; Yuan, Y.; Zuo, X.; Yue, Y.; Dai, W.; Fan, T.; Liu, G.; Liu, L.; et~al. 2025.
\newblock Dapo: An open-source llm reinforcement learning system at scale.
\newblock \emph{arXiv preprint arXiv:2503.14476}.

\bibitem[{Yuan et~al.(2024)Yuan, Pang, Cho, Sukhbaatar, Xu, and Weston}]{yuan2024self}
Yuan, W.; Pang, R.~Y.; Cho, K.; Sukhbaatar, S.; Xu, J.; and Weston, J. 2024.
\newblock Self-rewarding language models.
\newblock \emph{arXiv preprint arXiv:2401.10020}, 3.

\bibitem[{Yun et~al.(2025)Yun, Sohn, Park, Kim, Tang, Shao, Koo, Ko, Chen, Gerstein et~al.}]{yun2025med}
Yun, J.; Sohn, J.; Park, J.; Kim, H.; Tang, X.; Shao, Y.; Koo, Y.; Ko, M.; Chen, Q.; Gerstein, M.; et~al. 2025.
\newblock Med-PRM: Medical Reasoning Models with Stepwise, Guideline-verified Process Rewards.
\newblock \emph{arXiv preprint arXiv:2506.11474}.

\bibitem[{Zhang et~al.(2025{\natexlab{a}})Zhang, Lyu, Sun, Wang, Zhang, Hua, Wu, Guo, Wang, Muennighoff et~al.}]{zhang2025survey}
Zhang, Q.; Lyu, F.; Sun, Z.; Wang, L.; Zhang, W.; Hua, W.; Wu, H.; Guo, Z.; Wang, Y.; Muennighoff, N.; et~al. 2025{\natexlab{a}}.
\newblock A Survey on Test-Time Scaling in Large Language Models: What, How, Where, and How Well?
\newblock \emph{arXiv preprint arXiv:2503.24235}.

\bibitem[{Zhang et~al.(2025{\natexlab{b}})Zhang, Liu, Zhang, Liu, Luo, Huang, and Gong}]{zhang2025process}
Zhang, S.; Liu, X.; Zhang, X.; Liu, J.; Luo, Z.; Huang, S.; and Gong, Y. 2025{\natexlab{b}}.
\newblock Process-based self-rewarding language models.
\newblock \emph{arXiv preprint arXiv:2503.03746}.

\bibitem[{Zhang et~al.(2025{\natexlab{c}})Zhang, Sun, Zhang, Feng, Lu, Yang, and Meng}]{zhang2025critique}
Zhang, X.; Sun, H.; Zhang, Y.; Feng, K.; Lu, C.; Yang, C.; and Meng, H. 2025{\natexlab{c}}.
\newblock Critique-GRPO: Advancing LLM Reasoning with Natural Language and Numerical Feedback.
\newblock \emph{arXiv preprint arXiv:2506.03106}.

\bibitem[{Zheng et~al.(2023)Zheng, Chiang, Sheng, Zhuang, Wu, Zhuang, Lin, Li, Li, Xing et~al.}]{zheng2023judging}
Zheng, L.; Chiang, W.-L.; Sheng, Y.; Zhuang, S.; Wu, Z.; Zhuang, Y.; Lin, Z.; Li, Z.; Li, D.; Xing, E.; et~al. 2023.
\newblock Judging llm-as-a-judge with mt-bench and chatbot arena.
\newblock \emph{Advances in neural information processing systems}, 36: 46595--46623.

\end{thebibliography}

\end{document}